\def\year{2021}\relax
\documentclass[letterpaper]{article} 
\usepackage{aaai21}  
\usepackage{times}  
\usepackage{helvet} 
\usepackage{courier}  
\usepackage[hyphens]{url}  
\usepackage{graphicx} 
\usepackage{natbib}  
\usepackage{caption} 
\usepackage{xcolor}
\usepackage[switch]{lineno}

\usepackage{verbatim}

\title{Learning Accurate and Interpretable Decision Rule Sets from Neural Networks}

\author {
    Litao Qiao\thanks{The two authors contributed equally to this paper.},
    Weijia Wang\footnotemark[1],
    Bill Lin  \\
}
\affiliations {
    Electrical and Computer Engineering, University of California San Diego  \\
    l1qiao@eng.ucsd.edu, wweijia@eng.ucsd.edu, billlin@eng.ucsd.edu
}

\usepackage{multirow}
\usepackage{amsmath}
\usepackage{booktabs}
\usepackage{subcaption}
\usepackage{arydshln}

\begin{document}
\maketitle

\begin{abstract}
This paper proposes a new paradigm for learning a set of independent logical rules in disjunctive normal form as an interpretable model for classification.  We consider the problem of learning an interpretable decision rule set as training a neural network in a specific, yet very simple two-layer architecture.  Each neuron in the first layer directly maps to an interpretable if-then rule after training, and the output neuron in the second layer directly maps to a disjunction of the first-layer rules to form the decision rule set.  Our representation of neurons in this first rules layer enables us to encode both the positive and the negative association of features in a decision rule.  State-of-the-art neural net training approaches can be leveraged for learning highly accurate classification models.  Moreover, we propose a sparsity-based regularization approach to balance between classification accuracy and the simplicity of the derived rules.  Our experimental results show that our method can generate more accurate decision rule sets than other state-of-the-art rule-learning algorithms with better accuracy-simplicity trade-offs.  Further, when compared with uninterpretable black-box machine learning approaches such as random forests and full-precision deep neural networks, our approach can easily find interpretable decision rule sets that have comparable predictive performance.
\end{abstract}
\section{Introduction}
\label{sec:intro}

Machine learning is finding its way to impact every sector of our society, including healthcare, information technology, transportation, entertainment, business, and criminal justice.
In recent years, machine learning using neural networks have made tremendous advances in solving perceptual tasks like computer vision and natural language processing, with breakthrough performance in classification accuracy and generalization capability.
However, neural network methods have generally produced black box models that are difficult or impossible for humans to understand.
Their lack of interpretability makes it difficult to gain public trust for their use in high-stakes human-centered applications like medical-diagnosis and criminal justice, where decisions can have serious consequences on human lives
\cite{rudin-stop}.

Indeed, interpretability is a well-recognized goal in the machine learning community.
One popular approach to interpretable models is the use of decision rule sets \cite{ripper, rule-two, rule-ids, rule-brs, rule-cg}, where the model comprises an unordered set of independent logical rules in disjunctive normal form (DNF).
Decision rule sets are inherently interpretable because the rules are expressed in simple IF-THEN sentences that correspond to logical combinations of input conditions that must be satisfied for a classification.
An example of a decision rule set with three clauses is as follows:
\\
\\
\begin{tabular}{| l l |}
\hline
IF	& (age $\leq$ 50) OR \\
       	& (NOT smoker) OR \\
       	& (cholesterol $\leq$ 130 AND blood pressure $\leq$ 120) \\
THEN	& low heart disease risk. \\
\hline
\end{tabular}
\\
\\
In this example, the model would predict someone to have a low risk for heart disease if the person's cholesterol level and blood pressure are below the specified thresholds.
The model not only provides a prediction, but the corresponding matching rule also provides an explanation that humans can easily understand.
In particular, the explanations are stated directly in terms of the input features, which can be categorical (e.g., color equal to red, blue, or green) or numerical (e.g., age $\leq$ 50) attributes, where the binary encoding of categorical and numerical attributes is well-studied \cite{rule-brs, rule-cg}.

In this paper, we propose a new paradigm for learning accurate and interpretable decision rule sets as a neural network training problem.
In particular, we consider the problem of learning an interpretable decision rule set as training a neural network in a simple two-layer fully-connected neural network architecture called a Decision Rules Network (DR-Net).
In the first layer, called the Rules Layer, each trainable neuron with binary activation directly maps to a logical IF-THEN rule after training, where a positive input weight corresponds to a positive association of the input feature, a negative input weight corresponds to a negative association of the input feature, and a zero weight corresponds to an exclusion of the input feature.
In the second layer, called the OR Layer, the trainable output neuron with binary activation directly maps to a disjunction of the first-layer rules to form the decision rule set.

By formulating the interpretable rules learning problem as a neural net training problem, state-of-the-art training approaches (including recent advances) can be harnessed for learning highly accurate classification models, including well-developed stochastic gradient descent algorithms for effective training.
We are also able to leverage well-developed regularization concepts developed in the neural net community to trade off accuracy and model complexity in the training process.
In particular, we propose a sparsity-based regularization approach in which the model complexity in terms of the length of the rules and the number of rules are captured in a regularization loss function.
Minimizing the number of decision rules makes it easier for a user to understand all the conditions that correspond to a classification, and minimizing the lengths of the decision rules makes it easier for a user to interpret the explanations.
This regularization loss function can be combined with a binary cross-entropy loss function that measures training accuracy, so that the training process can balance between classification accuracy and the simplicity of the derived rule set.

Other benefits of a neural net based formulation is the availability of sophisticated development frameworks \cite{tensorflow, pytorch} for model development, powerful computing platforms (e.g., GPUs and deep learning accelerators) for efficient learning and inference, and other developments like federated learning \cite{fed-learning} that enables multiple entities to collaboratively learn a common, robust model without sharing data, which addresses critical data privacy and security concerns.

In comparison with previous rule-learning approaches, our approach has several notable advantages.
In \cite{rule-ids, rule-brs}, the pre-mining of frequent rule patterns is first used to produce a set of candidate rules,
from which various algorithmic approaches are used to select a set of rules from these candidates.
However, the requirement for pre-mining frequent rules limits the overall search space, thus hindering the algorithms from obtaining a globally optimized model. 
In \cite{rule-two, rule-cg}, the problem is formulated as an integer-programming problem in which the pre-mining of rules is not required, but approximations are required to solve large scale problems.
In contrast, our neural net based approach does not require rules mining and can take advantage of well-developed neural net training techniques to derive better interpretable models.
By connecting interpretable rule-based learning to a neural network based formulation, we hope to open a new line of research that will lead to further fruitful results in the future.

Our experimental results show that our method can generate more accurate decision rule sets than other state-of-the-art rule-learning algorithms with better accuracy-simplicity trade-offs.  Further, when compared with uninterpretable black box machine learning approaches such as random forests and full-precision deep neural networks, our approach can easily find interpretable decision rule sets that have comparable predictive performance.
\section{Decision Rules Network}
\label{sc:training}

Given a classification dataset with binarized input features, our goal is to train a classifier in the form of a Boolean logic function in disjunctive normal form (OR-of-ANDs). In particular, each of the lower level conjunctive clauses (logical ANDs), which consists of a subset of input features and their negations, individually serves as a decision rule. An instance satisfies a conjunctive clause if all conditions specified in the clause are true in the instance. In the upper level of the function, all conjunctive clauses are unified by a disjunction (logical OR). Thus, a negative final prediction is produced only if none of the conjunctive clauses are satisfied. Otherwise, a positive final prediction will be made.

Mathematically, the training set contains $N$ data samples $(\mathbf{x}_n, y_n)$, $n=1,...,N$, where $\mathbf{x}_n$ comprises $D$ binarized features $x_{n,i} \in \{0, 1\}$, $i=1,...,D$, and $y_n\in \{0, 1\}$. The final decision rule set $C$ learned from our method comprises parallel rules that we denote as clauses: $C = \{c_1, c_2, ... , c_m\}$. We define a clause $c$ to be a conjunction of $k$ predicates where $1 \leq k \leq D$ and a predicate to be either an input feature $x_i$ or the negation of an input feature $\overline{x}_i$. If an input feature or the negation of an input feature is not present in clause $c$, then we say that feature is excluded from clause $c$, i.e. whether $x_{n, i}$ is $0$ or $1$ has no effect to the prediction of clause $c$. Under this definition, an instance $\mathbf{x}_n$ satisfies a clause only if all predicates in the clause are true in the instance i.e. $x_{n, i} = 1$ for $x_i$ and $x_{n, i} = 0$ for $\overline{x}_i$. 

In this section, we introduce the architecture of our Decision Rules Network (DR-Net), which is a simple two-layer fully-connected neural network. The first layer, called the Rules Layer, consists of trainable neurons that map to logical IF-THEN rules, and the second layer, called the OR Layer, contains a trainable output neuron that maps to a disjunction of the first-layer rules to form the decision rule set. The goal of the design of this network is to simulate the logical formula in disjunctive normal form so that a trained DR-Net can be directly mapped to a set of interpretable decision rules.

\subsection{Handling of Categorical and Numerical Attributes}

Common tabular datasets generally comprise binary, categorical and numerical features. While our method is based on binary encoded input vectors, we employ the following pre-processing procedures, which are well established and studied in the machine learning literature, to binarize the input features. In particular, the values of binary features are left as what they are, whereas we apply standard one-hot encoding to transform categorical attributes to vectors of binary values. As for numerical features, we adopt quantile discretization to get a set of thresholds for each feature, where the original numerical value is one-hot-encoded into a binary vector by comparing with the thresholds (e.g., age $\leq$ 25, age $\leq$ 50, age $\leq$ 75) and encoded as $1$ if less than the threshold or $0$ otherwise. For example, considering a dataset that consists of the categorical feature ``color'' chosen from \{red, green, blue\} and a numerical feature ``age'' with thresholds \{25, 50, 75\}, our pre-processing approach will encode an instance [color: red, age: 30] as [red, green, blue, age $\leq$ 25, age $\leq$ 50, age $\leq$ 75] $=$ [1, 0, 0, 0, 1, 1]. Most other rule-learning methods \cite{rule-brs, rule-cg} require to convert binary, categorical and numerical features into both positive conditions, e.g., (color = blue) and (age $\leq$ 50), and negative conditions, e.g., (color $\neq$ blue) and (age $>$ 50), of the binary vectors in their pre-processing procedures. On the other hand, our encoding approach only involves those positive conditions without separately having their negations included. Further explanations will be discussed in the next section.

\subsection{Rules Layer}

The essence of a fully-connected layer is the dot-product operation shifted by a bias term. In this context, we notice that with binarized input features, a neuron can be constructed such that it effectively performs a logical AND operation by dynamically adjusting the bias based on the weight values and applying a binary step activation function afterwards. Then, by interpreting the full precision weights in a certain way, each neuron is effectively a conjunction of input features and thus the whole layer can be mapped to a set of clauses that can be later combined with disjunction to form a DNF rule set.

Mathematically, given the input to the Rules Layer as $\mathbf{x} \in \{0, 1\}^D$ and the output as $y$, a neuron in the Rules Layer performs its operation as follows:
\begin{equation}
\label{eq:rule-neuron}
    y = \sum_{i=0}^D w_ix_i - \sum_{w_i > 0}w_i + 1.
\end{equation}
In Equation~\ref{eq:rule-neuron}, the dot product of the weights and inputs is added with a dynamic bias, which depends on the weights of the neuron. With the dynamic bias and binarized inputs, the range of the outputs of the neurons in the Rules Layer is within $(-\infty, 1]$. Note that the output $y=1$ can only be achieved when all inputs match the sign of the corresponding weights: all positive weights should have the inputs of $1$ and all negative weights should have the inputs of $0$. Just like the behavior of weights in regular neurons, the zero weights in the Rules Layer mean that the corresponding inputs will not have any effect on the output.

In order for the neuron in the Rules Layer to function as a proper logical AND operation, we need to apply a binary step activation function to its output:
\begin{equation}
\label{eq:rule-activation}
    f(x) = 
    \begin{cases}
    1 & \text{if } x = 1 \\
    0 & \text{otherwise}
    \end{cases}
\end{equation}
When applied at the Rules Layer, the binary step function defined in Equation~\ref{eq:rule-activation} simply maps the range $(-\infty, 1)$ to 0, which ensures that the neuron is turned on only when Equation~\ref{eq:rule-neuron} evaluates to 1. With the dynamic bias and binary step function, each neuron in the Rules Layer encodes a rule that has $k$ predicates, where $k$ is the number of non-zero weights of that neuron. As discussed earlier, in effect neuron in the Rules Layer maps to a logical IF-THEN rule after training, where a positive input weight corresponds to a positive association of the input feature, a negative input weight corresponds to a negative association of the input feature, and a zero weight corresponds to an exclusion of the input feature.

However, as can be observed, the activations of the first layer are discretized into binary integers that are not naturally differentiable and the classic gradient computation approach doesn't apply here. Therefore, we utilize the straight-through estimator discussed in \cite{estimator} with the gradient clipping technique. Denoted by $\hat{y}_i$ the binarized activation based on $y_i$, we compute the gradient as follows:
\begin{equation}
    g_{\hat{y}_i} = \Bigg\{
    \begin{array}{@{}ll@{}}
         \multirow{2}{*}{$0$} & \textrm{if }y_i < 0 \\
          & \textrm{or } y_i > 1 \frac{\partial L}{\partial y_i} < 0 \\
        g_{y_i} &\textrm{otherwise}
    \end{array}
    \label{eq:grad}
\end{equation}
where $g_{\hat{y}_i}$ and $g_{y_i}$ are the gradients of classification loss w.r.t. $\hat{y}_i$ and $y_i$, respectively. The condition $y_i < 0$ simulates the backward computation of the ReLU function, which introduces non-linearity into the training process and empirically improves the performance; whereas our motivation of the second condition is to address the saturation effect: we suppress the update of the full-precision activations that are greater than $1$ and are still driven by the gradient to increase, since further raising activations does not produce any difference after binarization.

As discussed in above, the addition of the negative conditions in the input space is critical to the selection-based methods \cite{rule-brs, rule-cg} since they only consider the presence and absence of features and cannot deduce negative correlations unless they are explicitly provided in the input space. On the other hand, besides the presence of a positive association or an exclusion, our Rules Layer also learns the negation of an input feature by assigning a negative weight to it, and hence, DR-Net can directly derive negative conditions from the corresponding input features. Therefore, appending negative conditions in the input binary vector is redundant in DR-Net, and the input space of our DR-Net is reduced by half comparing with those selection-based method.

\subsection{OR Layer}

\begin{figure*}[t]
\begin{center}
\includegraphics[width=0.95\textwidth]{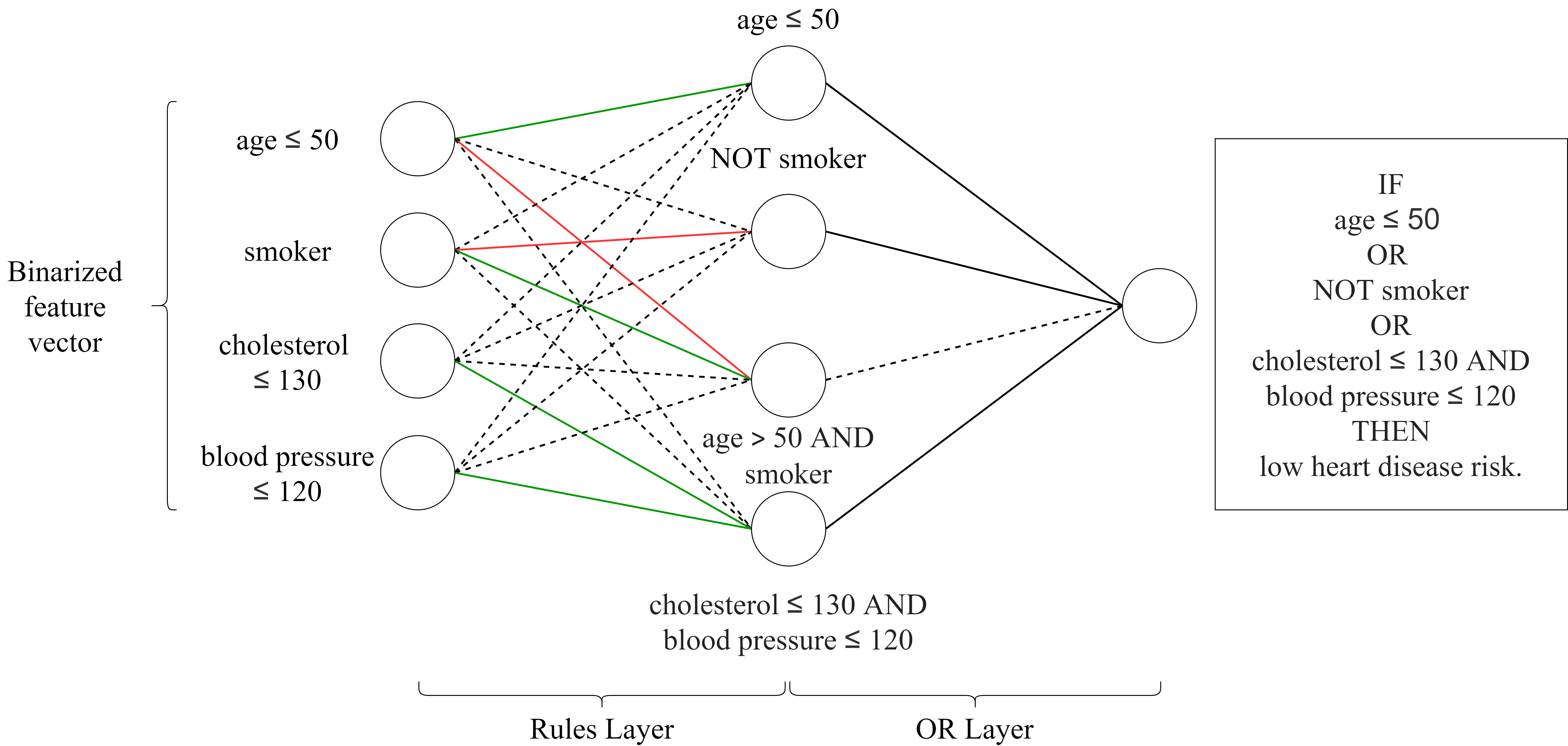}
\caption{An example of the DR-Net architecture where three rules from the Rules Layer are included in the OR Layer, and one rule is excluded. The decision rule set that the network directly maps into is shown in the box on the right. The dashed lines represent the masked weights (weights that are set to zero). The green lines in the Rules Layer represent positive weights while red lines represent negative weights. Please note that we represent (NOT age $\leq$ 50) as (age $>$ 50) in the third rule, and it is not included in the final rule set because it has been masked in the OR Layer.}
\label{fig:arch}
\end{center}
\end{figure*}

To produce the disjunction of the logical rules learned in the Rules Layer, the OR Layer contains only one output neuron, where the weights are binarized as follows:
\begin{equation}
    \hat{w_i} = \Bigg\{
    \begin{array}{@{}ll@{}}
        0 & \textrm{if }w_i \leq 0 \\
        1 &\textrm{otherwise}
    \end{array}
    \label{eq:binarize}
\end{equation}
The output neuron performs a dot product with a negative bias $-\epsilon$ as follows:
\begin{equation}
    y = \sum_{i=1}^D \hat{w_i}x_i - \epsilon,
\end{equation}
where $0 < \epsilon < 1$ is a small value such that $y$ is positive when at least one input is activated. With a sigmoid activation function and a binary cross-entropy loss, this particular neuron behaves as an OR gate: the output is by default turned off because of the negative bias, while it produces a positive value if at least one rule is activated with a corresponding $\hat{w_i} = 1$, which exactly mimics the behavior of the logical OR function. The binarized weights $\hat{w_i}$ act as rule selectors that filter out rules that do not contribute to the model's predictive performance. An example of our complete network structure is shown in Figure~\ref{fig:arch}. We practically use $\epsilon=0.5$ in our implementations.

\subsection{Sparsity-Based Regularization}

The neural network structure proposed above outlines a way to derive a set of decision rules using stochastic gradient descent. As discussed above, a zero weight for a Rules Layer neuron corresponds to the exclusion of the corresponding input feature. Similarly, a zero weight for the OR Layer output neuron corresponds to the exclusion of the corresponding rule from the rule set. Thus, it should be clear that maximizing the \emph{sparsity} of the Rules Layer neurons corresponds to simplifying the corresponding rules, and maximizing the sparsity of the OR Layer neuron corresponds to minimizing the number of rules.

However, to eliminate an input feature from a logical rule or a logical rule from the complete rule set, the corresponding weight has to be exactly zero, which is difficult to achieve in the typical network training process. To achieve a high degree of sparsity with exact zero weights, we explicitly incorporate a \emph{sparsity-based regularization} mechanism into the training process using an approach akin to $L_0$ regularization by explicitly training \emph{mask} variables.

As discussed in \cite{l0} as a way to achieve network sparsity through $L_0$ regularization, a binary random variable $z_i \in \{0, 1\}$ is attached to each weight of the model to indicate whether the corresponding weight is kept or removed. With this, we can reparameterize each weight $w_i$ as the product of a weight $\Tilde{w_i}$ and the corresponding binary random variable $z_i$:
\begin{equation}
    w_i = \Tilde{w}_i z_i.
\end{equation}
Assuming each $z_i$ is subject to a Bernoulli distribution with parameter $\pi_i$, i.e. $q(z_i|\pi_i) = \textrm{Bern}(\pi_i)$, the probability that $z_i$ is 1 is just $\pi_i$. In \cite{l0}, $L_0$ regularization is implemented by summing all $\pi_i$ parameters as the penalty term in the loss function\footnote{As explained below, we do not use $L_0$ normalization as the regularization term. Instead, we explicitly capture the model complexity with Equation~\ref{eq:reg} as the regularization term.}. In order to train the binary random variables with stochastic gradient descent, two different gradient estimators have been proposed in \cite{l0} and \cite{l0arm}, respectively, to approximate the Bernoulli distribution.

Applying the above regularization method to the Rules Layer is straightforward: all weight parameters are replaced by their product with the corresponding mask variables. For the OR Layer, since the weights will ultimately be binarized, we can just directly substitute the mask variables for the weights to simplify the process. That is, we can simply treat $w_i = z_i$, with no need for a separate $\Tilde{w_i}$ variable\footnote{Since $w_i = z_i$ is already binarized, there is no need to further binarize $w_i$ to derive $\hat{w_i}$ with Equation \ref{eq:binarize}. i.e., we can just use $\hat{w_i} = w_i = z_i$.}.

We then incorporate a sparsity-based regularization term in the loss function to model the complexity of the rule set represented by the neural network. We denote by $\pi_{1,i,j}$ and $\pi_{2,j}$ the penalty of the non-zero mask variables of the Rules Layer and the OR Layer, respectively, where $i = 1,2,\dots,D$ is the feature index, and $j = 1,2,\dots,m$ is the index to the $j$-th neuron (rule). Then the regularization loss is defined as follows:
\begin{equation}
\label{eq:reg}
    \mathcal{L}_R = \frac{1}{m}\left(\sum_{j=1}^m \pi_{2,j}  + \sum_{j=1}^m \pi_{2,j} \sum_{i=1}^D \pi_{1,i,j}\right),
\end{equation}
which \emph{explicitly} captures the model complexity, as similarly defined in \cite{rule-cg}. In particular, the model complexity of a rule set is defined as the sum of the number of rules and the total number of predicates in all rules. Following this definition, the first and the second terms of Equation~\ref{eq:reg} quantify the losses for the number of rules and the total number of predicates, respectively. Note that, according to the second term in Equation~\ref{eq:reg}, the loss for the number of predicates in the $j$-th rule will be effectively removed if $\pi_{2,j}$ is at or near zero: i.e., the $j$-th neuron in the Rules Layer is disconnected from the OR Layer.

With the above sparsity-based regularization applied to DR-Net, the overall loss function we optimize for can be expressed as follows:
\begin{equation}
\label{eq:loss}
    \mathcal{L} = \mathcal{L}_\mathrm{BCE} + \lambda \mathcal{L}_{R},
\end{equation}
where $\mathcal{L}_\mathrm{BCE}$ is the binary cross-entropy loss, $\mathcal{L}_{R}$ is the regularization penalty that is specified by Equation~\ref{eq:reg}, and $\lambda$ is the regularization coefficient that balances the classification accuracy and rule set complexity.

\subsection{Alternating Two-Phase Training Strategy}

As previously discussed, each neuron in the first layer (the Rules Layer) of our proposed network architecture encodes an interpretable decision rule, whereas the output neuron in the second layer (the OR Layer) chooses some of the rules to be included in the set of decision rules. Empirically, we noticed that it is more effective to train our DR-Net with gradient-based optimizers (e.g., SGD) in an alternating manner, potentially due to the reduced search space and simpler optimization goals. In particular, our ``alternating training strategy'' consists of two training phases. We first freeze the OR Layer and only update the parameters in the Rules Layer to learn plausible rules. In the second phase, the Rules Layer is then fixed and we optimize the OR Layer such that redundant rules are eliminated while necessary inactive rules can also be re-enabled. The whole network is trained by alternating between the two training phases until convergence.

In addition, since the sparsity of the Rules Layer is directly related to the simplicity of rules, whereas the second layer is more focused on the selection of these derived rules, we further allow the flexible weighting of the sparsity-based regularization loss of the two layers. Specifically, as illustrated in Equation~\ref{eq:loss}, the balance between classification loss and regularization loss is implemented via the regularization coefficient $\lambda$, where we can practically use different values for the two phases. In other words, the Rules Layer and the OR Layer are optimized over $\mathcal{L}_1$ and $\mathcal{L}_2$, respectively, where $\mathcal{L}_1$ and $\mathcal{L}_2$ are defined as follows:
\begin{equation}
\label{eq:alter}
    \begin{array}{c}
         \mathcal{L}_1 = \mathcal{L}_\mathrm{BCE} + \lambda_1 \mathcal{L}_{R}, \\
         \mathcal{L}_2 = \mathcal{L}_\mathrm{BCE} + \lambda_2 \mathcal{L}_{R}.
    \end{array}
\end{equation}
In this way, the trade-off between model simplicity and accuracy in our experiments can be modulated by the adjustments of $\lambda_1$ and $\lambda_2$.
\section{Experimental Evaluation}
\label{sec:exp}

The numerical experiments were evaluated on 4 publicly available binary classification datasets, which all have more than 10,000 instances and more than 10 attributes for each instance before binarization. The first two selected datasets are from UCI Machine Learning Repository \cite{uci}: MAGIC gamma telescope (magic) and adult census (adult), which are also used in recent works on rule set classifiers \cite{rule-cs, rule-brs, rule-cg}. The magic dataset is a dataset with pure numerical attributes while the adult dataset has a mix of both categorical and numerical attributes. The other two datasets are relatively recent datasets: the FICO HELOC dataset (heloc) and the home price prediction dataset (house), which have all numerical attributes. 
In all datasets, pre-processing is performed to encode categorical and numerical attributes into binary variables, as discussed earlier in the paper. Also, we append negative conditions for all other models except DR-Net.

Our goal is to learn a set of decision rules using our DR-Net and compare our model with other state-of-the-art rule learners and machine learning models. The results include model accuracies and complexities. Apart from the model complexity defined earlier (the number of rules plus the total number of conditions in the rule set), we also define the rule complexity, which is the average number of conditions in each rule of the model. We consider three other rule learners to directly compare with our work in terms of both accuracy and interpretability: the RIPPER algorithm \cite{ripper}, Bayesian Rule Sets (BRS) \cite{rule-brs}, and the Column Generation (CG) algorithm from \cite{rule-cg}. The first one is an old rule set learning algorithm that is a variant of the Sequential Covering algorithm, while the other two are representatives of recent works in rule learning classifiers. We used open-source implementations on GitHub for all three algorithms, where the CG implementation \cite{aix360} is slightly modified from the original paper. Other models used for comparison are the scikit-learn \cite{sklearn} implementations of the decision tree learner CART \cite{cart} and Random Forests (RF) \cite{rf}. We also include a full-precision deep neural network (DNN) model with 6 layers, 50 neurons per hidden layer and ReLU activations. The last two models are \emph{uninterpretable} models intended to provide baselines for typical performances that black-box models can achieve on these datasets. These uninterpretable baseline results serve as benchmarks for accuracy comparisons.

For DR-Net, we used the Adam optimizer with a fixed learning rate of $10^{-2}$ and no weight decay across all experiments. There are 50 neurons in the Rules layer to ensure there is an efficient search space for all datasets. The alternating two-phase training strategy discussed earlier is employed with 10,000 total number of training epochs and 1,000 epochs for each layer. For simplicity, the batch size is fixed at 2,000 and the weights are uniformly initialized within the range between $0$ and $1$. The parameters that are related to sparsity-based regularization are set the same as in the original paper \cite{l0}. 

\subsection{Classification Performance}

We evaluated the predictive performance of DR-Net by comparing both test accuracy and complexity with other state-of-the-art machine learning models. 5-fold nested cross validation was employed to select the parameters for all rule learners that explicitly trade-off between accuracy and interpretability to maximize the training set accuracies. To ensure that the final rule learner models are interpretable, we constrained the possible parameters for nested cross validation to a range that results in a low model complexity. For DR-Net, We fixed the $\lambda_2$ to be $10^{-5}$ in Equation~\ref{eq:alter} and only $\lambda_1$ was varied in the experiment. Although there are many parameters in BRS to control the rule complexity, we followed the procedure used in \cite{rule-cg} and only varied the multiplier $\kappa$ in prior hyper-parameter to save running time. For RIPPER, we varied the maximum number of conditions and the maximum number of rules as hyper-parameters of the implementation, which are directly related to the complexity of the model. The CG implementation in \cite{aix360} doesn't have the complexity bound parameter $C$ as specified in \cite{rule-cg} but instead provides two hyper-parameters to specify the costs of each clause and of each condition, which were used in our experiment to control the rule set complexity. We left all other parameters for these three algorithms (CG, BRS, RIPPER) as default. For CART and RF, we constrained the maximum depth of trees to be 100 for all datasets to achieve better generalization. For DNN, we used the same training parameters (number of epochs, batch size, learning rate, etc.) with a weight decay of $10^{-2}$. The test accuracy results of all models on all datasets are shown in Table~\ref{tab:accuracy} and the corresponding complexities are shown in Table~\ref{tab:complexity}. We omitted the results of the complexities of CART, RF and DNN because they have a different notion of model complexity and rule complexity.

\begin{table}[htbp]
\begin{center}
\begin{tabular}{ccccc}
\toprule
dataset                 & magic         & adult         & heloc         & house             \\
\midrule
\midrule
\multicolumn{5}{c}{interpretable} \\
\midrule
\multirow{2}{*}{DR-Net} & \textbf{84.42}& \textbf{82.97}& \textbf{69.71}& \textbf{85.71}    \\
                        & (0.53)        & (0.51)        & (1.05)        & (0.40)            \\
\midrule
\multirow{2}{*}{CG}     & 83.68         & 82.67         & 68.65         & 83.90             \\
                        & (0.87)        & (0.48)        & (3.48)        & (0.18)            \\
\midrule
\multirow{2}{*}{BRS}    & 81.44         & 79.35         & 69.42         & 83.04             \\
                        & (0.61)        & (1.78)        & (3.72)        & (0.11)            \\
\midrule
\multirow{2}{*}{RIPPER} & 82.22         & 81.67         & 69.67         & 82.47             \\
                        & (0.51)        & (1.05)        & (2.09)        & (1.84)            \\
\midrule
\multirow{2}{*}{CART}   & 80.56         & 78.87         & 60.61         & 82.37             \\
                        & (0.86)        & (0.12)        & (2.83)        & (0.29)            \\
\midrule
\midrule
\multicolumn{5}{c}{uninterpretable} \\
\midrule
\multirow{2}{*}{RF}     & 86.47         & 82.64         & 70.30         & 88.70             \\
                        & (0.54)        & (0.49)        & (3.70)        & (0.28)            \\
\midrule
\multirow{2}{*}{DNN}    & 87.07         & 84.33         & 70.64         & 88.84             \\
                        & (0.71)        & (0.42)        & (3.37)        & (0.26)            \\
\bottomrule
\end{tabular}
\end{center}
\caption{Test accuracy based on the nested 5-fold cross validation (\%, standard error in parentheses).}
\label{tab:accuracy}
\end{table}

It can be seen in Table~\ref{tab:accuracy} that our method outperforms other interpretable models on all datasets. For these better accuracy results, our method does not establish a similar superiority in the complexity comparison (Table~\ref{tab:complexity}). However, as shown Figure~\ref{fig:trade-off} and further discussed in the next section, our DR-Net approach can often achieve higher accuracy at comparable complexities. It is interesting to see that DR-Net maintains a relatively good model complexity compared with the corresponding rule complexity, which is exactly because our regularization loss function is designed specifically to minimize the model complexity instead of the rule complexity. Compared with RIPPER, which greedily mines good rules in each iteration to maximize the training accuracy, DR-Net is very competitive in the sense that it has similar or better test performance while consistently maintaining a lower model complexity. One advantage of the BRS algorithm over other models is that it consistently generates sparse models across all datasets, but at the expense of significantly inferior accuracies. The CART decision tree algorithm turned out to be the worst performing model in our experiments, which might result from overfitting. The results in Table~\ref{tab:accuracy} and Table~\ref{tab:complexity} suggest that our DR-Net approach is very competitive as a machine learning model for interpretable classification. Finally, our DR-Net approach is able to achieve accuracies within only 3\% of the uninterpretable models (RF and DNN) on the datasets evaluated. 

\begin{table}[htbp]
\begin{center}
\begin{tabular}{ccccc}
\toprule
dataset                  & magic         & adult         & heloc         & house         \\
\midrule
\midrule
\multirow{2}{*}{DR-Net}  & 109.4         & 86.0          & 13.8          & 85.0          \\
                         & 5.22          & 13.54         & 6.33          & 6.31          \\
\midrule
\multirow{2}{*}{CG}      & 112.8         & 120.0         & \textbf{3.4}  & \textbf{28.6} \\
                         & 3.72          & 3.77          & \textbf{1.90} & 5.15          \\
\midrule
\multirow{2}{*}{BRS}     & \textbf{40.0} & \textbf{16.8} & 16.6          & 31.2          \\
                         & \textbf{3.00} & \textbf{3.00} & 2.96          & \textbf{3.00} \\
\midrule
\multirow{2}{*}{RIPPER}  & 189.4         & 117.6         & 72.8          & 328.0         \\
                         & 6.01          & 4.66          & 5.24          & 7.01          \\
\bottomrule
\end{tabular}
\end{center}
\caption{Model complexity (upper) and rule complexity (lower) corresponding to the accuracy results shown in Table~\ref{tab:accuracy} based on the nested 5-fold cross validation. While DR-Net, using parameters selected by the nested 5-fold cross validation with the priority for accuracy, does not achieve the best complexity in comparison with other models, it can be observed in Figure~\ref{fig:trade-off} that our approach can generally achieve a higher accuracy at the cost of comparable complexities.}
\label{tab:complexity}
\end{table}

\subsection{Accuracy-Complexity Trade-off}
\label{sec:trade-off}

\begin{figure*}[th!]
\begin{center}
\begin{subfigure}[b]{\textwidth}
\includegraphics[width=\textwidth]{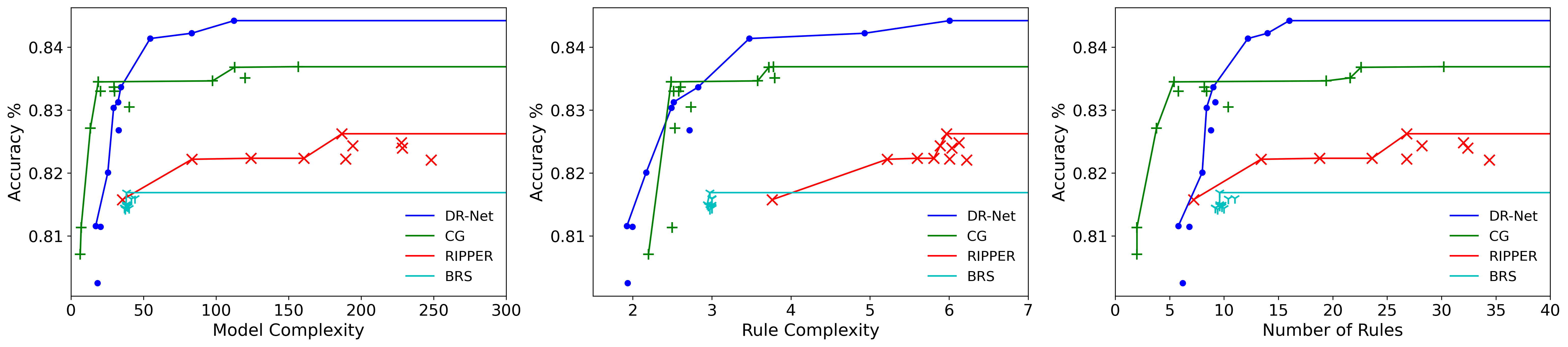}
\caption{magic}
\end{subfigure}
\begin{subfigure}{\textwidth}
\includegraphics[width=\textwidth]{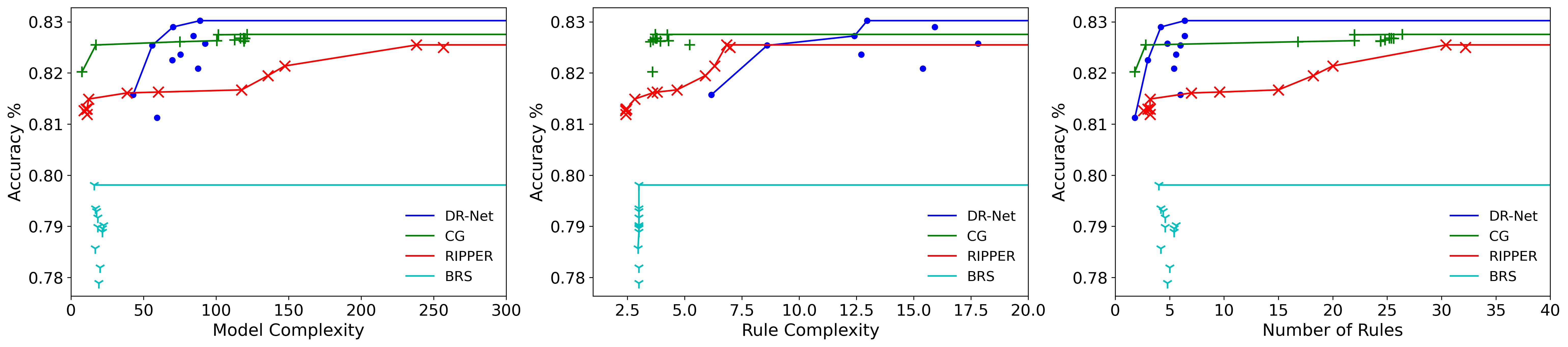}
\caption{adult}
\end{subfigure}
\begin{subfigure}{\textwidth}
\includegraphics[width=\linewidth]{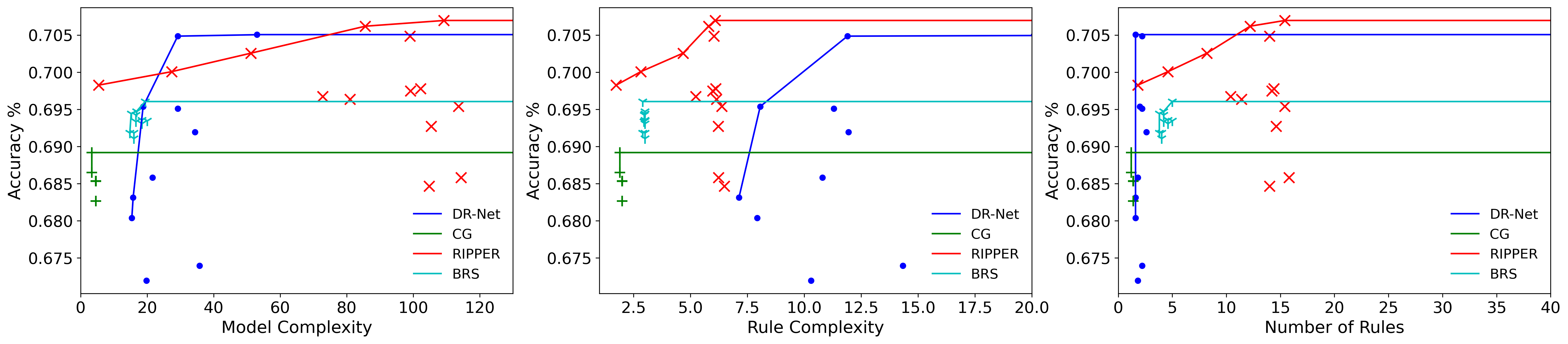}
\caption{heloc}
\end{subfigure}
\begin{subfigure}{\textwidth}
\includegraphics[width=\linewidth]{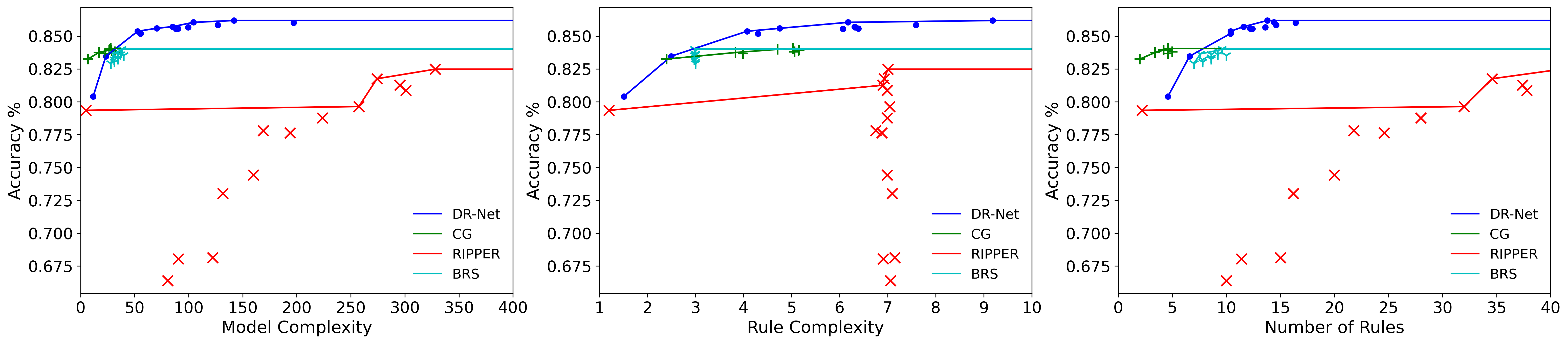}
\caption{house}
\end{subfigure}
\caption{Accuracy-Complexity trade-offs on all datasets. Pareto efficient points are connected by line segments.}
\label{fig:trade-off}
\end{center}
\end{figure*}

In this experiment, we compared the accuracy-complexity trade-off of our DR-Net with other rule learning algorithms: CG, BRS and RIPPER. The parameters that were selected to be varied in this experiment are the same as ones in the first experiment. Instead of using nested cross validation to select best parameters on the validation set, we manually picked a set of values for each selected parameters for each algorithm to generate different sets of accuracy-complexity pairs. We ran the experiments on all datasets and the results with the average of the 5-fold cross validation are shown in Figure~\ref{fig:trade-off}. Apart from model complexity and rule complexity, we included a third metric to show the average number of rules in each generated rule set versus the test accuracy. For each method compared, the dots connected by the line segments shown correspond to Pareto efficient models where all other points below the Pareto frontier have either lower accuracies or higher complexities.

The characteristic of being able to attain a high test accuracy with an acceptable model complexity for DR-Net in Table~\ref{tab:accuracy} and Table~\ref{tab:complexity} is carried over to Figure~\ref{fig:trade-off}. For the magic, adult and house datasets, DR-Net outperforms all other rule learners in terms of the accuracy by a substantial margin when the model complexity, the rule complexity or the number of rules exceeds a certain threshold. Although DR-Net does not dominate RIPPER on the heloc dataset, their accuracy comparison is very close if enough model complexity or number of rules is given. The only thing that DR-Net falls behind a little bit is in the rule complexity vs. accuracy comparison on the heloc dataset. In theory, DR-Net can achieve relatively low rule complexity with a different regularization loss function that can quantify the average number of conditions in the rule set, which we leave as future work. It is also interesting to note that the number of rules from DR-Net varies in a relatively narrower range compared with other approaches as shown in the third column of Figure~\ref{fig:trade-off}, which is directly resulted by fixing $\lambda_2$ in Equation~\ref{eq:alter}. BRS does not demonstrate a clear accuracy-complexity trade-off as its results all group in a very narrow range, which is also noted and explained in \cite{rule-cg}. This experiment shows that DR-Net can be preferred over other rule learners because of its potential for achieving a much higher test accuracy with a relatively moderate complexity sacrifice.
\section{Related Work}
\label{sec:related}

The learning of Boolean rules and rule sets is well studied with different variants.
While the learning of two-level Boolean decision rule set has an extensive history in different communities, most of them employ heuristic algorithms that optimize for certain criteria that are not directly related to classification accuracy or model simplicity.
Representatives of these methods include logical analysis of data \cite{LAD2, LAD1}, association rule mining and classification \cite{CN2, CBA}, and greedy set covering \cite{ripper}.

With the increasing interest in the field of explainable machine learning, researchers have  in recent years added model complexity to the optimization objective so that accuracy and simplicity can be jointly optimized.
Several approaches select rules from a pre-mined set of candidate rules \cite{rule-brs, rule-ids}.
A Bayesian framework is presented in \cite{rule-brs} for selecting pre-mined rules by approximately constructing a maximum a posteriori (MAP) solution.
In \cite{rule-ids}, the joint optimization problem is approximately solved by a local search algorithm.
In these methods, the requirement for rules pre-mining limits the overall search space, hindering their ability to find a globally optimized model.
Other approaches based on integer-programming (IP) formulations \cite{rule-two, rule-cg} do not require rules pre-mining, but they rely on approximate solutions for large datasets.
In \cite{rule-cg}, the IP problem is approximately solved by relaxing it into a linear programming problem and applying the column generation algorithm, whereas \cite{rule-two} utilizes various optimization approaches including block coordinate descent and alternating minimization algorithm.
 
Besides decision rule sets, decision lists \cite{rivest87, bertsimas11, brl} and decision trees \cite{cart, dt} are also interpretable rule-based models.
In decision lists, rules are ordered in an IF-THEN-ELSE sequence.
However, the chaining of rules via an IF-THEN-ELSE sequence means that the interpretation of an activated rule requires an understanding of all preceding rules.
This can make the explanation more difficult for humans to understand.
In decision trees, rules are organized into a tree structure.
However, they are often prone to overfitting.
\section{Conclusion and Extensions}

In this paper, we presented a simple two-layer neural network architecture, which can be directly mapped to a set of interpretable decision rules, along with a procedure to accurately train the network for classification. 
We described a sparsity-based regularization approach that can capture the complexity of the trained model in terms of the length of the rules and the number of rules.
The incorporation of this regularization loss into the overall loss function enables the training process to balance between classification accuracy and model complexity.
With our neural net formulation, we are able to leverage state-of-the-art neural net infrastructures to learn highly accurate and interpretable rule-based models.
Our experimental results show that our method can generate more accurate decision rule sets than other state-of-the-art rule-learners with better accuracy-simplicity trade-offs.
When compared with uninterpretable black box models such as random forests and full-precision deep neural networks, our approach can easily learn interpretable models that have comparable predictive performance.

We focus in this paper on the binary classification problem, but the approach can be easily extended to multi-class classification by deploying separate output neurons for each class and mapping each output neuron to a corresponding set of rules for the respective class.
A default class and a tie-breaking function could be used in the event that no class or more than one class is activated, respectively \cite{rule-ids},
or these cases can be handled by error correcting output codes \cite{schapire97}.
We plan to investigate in future work potentially more powerful tie-breaking mechanisms that can be directly trained as part of the neural net formulation, for example by directly interpreting softmax results.

\section*{Acknowledgements}

This research has been supported in part by the National Science Foundation (NSF IIS-1956339).

\bibliography{references}

\end{document}